\documentclass[conference]{IEEEtran}
\IEEEoverridecommandlockouts

\usepackage{cite}
\usepackage{amsmath,amssymb,amsfonts}
\usepackage{algorithmic}
\usepackage{graphicx}
\usepackage{textcomp}
\usepackage{xcolor}
\usepackage{multirow}
\usepackage{booktabs}
\usepackage{stfloats}
\graphicspath{{./}}

\def\BibTeX{{\rm B\kern-.05em{\sc i\kern-.025em b}\kern-.08em
    T\kern-.1667em\lower.7ex\hbox{E}\kern-.125emX}}

\begin{document}

\title{CogEvo-Edu: Cognitive Evolution Educational Multi-Agent Collaborative System}

%
%
%
%
%

\author{
	Yefeng Wu\textsuperscript{1}\textsuperscript{,}\textsuperscript{*}, Yuchen Song\textsuperscript{2}, Yecheng Zhao\textsuperscript{1}, Ling Wu\textsuperscript{1}, Shan Wan\textsuperscript{1} \\
	\textsuperscript{1}Electronic Science and Technology, Anhui University, Hefei, China \\
	\textsuperscript{2}Medical Imaging Science, Wannan Medical College, Wuhu, China \\
	\textsuperscript{*}Emails: wuyefengflc@163.com
}

\maketitle

\begin{abstract}
Large language models (LLMs) are increasingly deployed as conversational tutors in STEM education, yet most systems still rely on a single LLM with a static retrieval‑augmented generation (RAG) pipeline over course materials. This design struggles in complex domains such as digital signal processing (DSP), where tutors must maintain coherent long‑term student models, manage heterogeneous knowledge bases, and adapt teaching strategies over extended interactions. We argue that retrieval, memory, and control should be treated as a coupled cognitive evolution process. We instantiate this view in CogEvo‑Edu, a hierarchical educational multi‑agent system comprising a Cognitive Perception Layer (CPL), a Knowledge Evolution Layer (KEL), and a Meta‑Control Layer (MCL). CPL maintains dual memories and performs confidence‑weighted consolidation to build structured, self‑correcting student profiles under limited context. KEL assigns each knowledge chunk a spatiotemporal value that drives activation, semantic compression, and forgetting. MCL formulates tutoring as hierarchical sequential decision making, orchestrating specialized agents and jointly adapting CPL/KEL hyperparameters via a dual inner–outer loop. To evaluate CogEvo‑Edu, we construct DSP‑EduBench, a vertical benchmark for DSP tutoring with heterogeneous resources, simulated student profiles, and long‑horizon interaction scripts. Using a three‑model LLM‑as‑a‑Judge ensemble, CogEvo‑Edu raises the overall score from 5.32 to 9.23 and improves all six indicators over static RAG, simple memory, and a single‑agent variant, demonstrating the value of jointly evolving student profiles, knowledge bases, and teaching policies.
\end{abstract}


\section{INTRODUCTION}

Large language models (LLMs) are increasingly used as conversational tutors in STEM education, where they can explain concepts, diagnose misconceptions, and generate practice problems\cite{nye2023generative,liu2024advancing}. Most deployed systems, however, still follow a simple pattern: a single LLM front-end plus a static retrieval-augmented generation (RAG) pipeline over course materials\cite{lewis2020retrieval,liu2025lpitutor}. In complex domains such as digital signal processing (DSP), this design struggles to maintain coherent long-term student models, manage knowledge bases efficiently, and adapt teaching strategies over time.

Three limitations are particularly critical. First, with finite context windows, most tutors approximate student state using sliding windows or coarse summaries, which leads to catastrophic forgetting and inconsistent personalization in long dialogues\cite{zhong2024memorybank,packer2023memgpt,maharana2024evaluating}. Second, the knowledge base is typically a fixed vector store queried with hand-tuned top-$k$ retrieval, which causes redundant ``retrieval piling'' and ignores how the pedagogical value of knowledge chunks changes with use and recency. Recent dynamic retrieval-augmented generation methods aim to decide when and what to retrieve during generation\cite{su2024dragin}, but they still optimize generic knowledge-intensive tasks rather than long-horizon tutoring. Third, teaching control is mostly monolithic: one LLM is expected to handle diagnosis, explanation, questioning, and retrieval, sometimes aided by ad-hoc rules, leaving little room for systematic optimization of ``what to teach, how to teach, and who teaches'' as data accumulates; yet multi-agent LLM frameworks in other domains demonstrate that explicitly orchestrated specialists can improve complex tool use and task decomposition\cite{wu2023autogen,li2023camel}.

We address these via \emph{cognitive evolution}, treating retrieval, memory, and control as coupled processes in CogEvo-Edu, a hierarchical educational multi-agent system. The Cognitive Perception Layer (CPL) maintains dual memories with confidence-weighted consolidation for self-correcting student profiles. The Knowledge Evolution Layer (KEL) assigns spatiotemporal value scores (interaction frequency, temporal decay, semantic density) to drive activation, compression, and forgetting. The Meta-Control Layer (MCL) orchestrates specialized agents and adapts CPL/KEL hyperparameters via dual inner--outer loops.

We evaluate on DSP-EduBench, a vertical benchmark with heterogeneous resources, simulated profiles, and long-horizon scripts. A three-model LLM-as-a-Judge ensemble scores six indicators. CogEvo-Edu raises the average from $5.32$ (vanilla LLM) to $9.23$, outperforming static RAG, simple memory, and single-agent variants across all dimensions. Contributions: (i) the cognitive evolution architecture; (ii) CPL/KEL mechanisms with value-based consolidation and lifecycle management; (iii) DSP-EduBench with multi-dimensional LLM-as-a-Judge evaluation.

\section{RELATED WORK}

\subsection{LLM-based Educational Systems and RAG Tutors}

Traditional intelligent tutoring systems (ITS) relied on explicit domain models and rule-based dialogue managers\cite{corbett1994knowledge,lord1968statistical}. Recent systems replace these components with general-purpose LLMs, building generative tutors that can explain concepts, answer questions, and produce exercises\cite{nye2023generative,liu2024advancing}. Many integrate retrieval-augmented generation (RAG)\cite{lewis2020retrieval,liu2025lpitutor} over textbooks or lecture notes to reduce hallucinations. Yet the knowledge store is usually static and retrieval configurations are fixed, so the system cannot adapt its knowledge structure to usage patterns or student needs. A single LLM still performs diagnosis, explanation, and retrieval inside one monolithic model. CogEvo-Edu instead decomposes these responsibilities across agents and layers and treats the knowledge base as an evolving object controlled by a value function.

\subsection{Student Modeling, Memory, and Retrieval}

Student modeling has long been studied in ITS through model tracing and knowledge tracing methods that maintain latent mastery variables for predefined skills\cite{corbett1994knowledge,lord1968statistical}, but these techniques are designed for discrete item-response data rather than open-ended dialogue. In current LLM-based tutors, student state is typically approximated with long context windows, sliding histories, or coarse dialogue summaries, sometimes plus simple user profiles. Several recent architectures instead introduce explicit short-term and long-term memory modules for LLM agents\cite{zhong2024memorybank,packer2023memgpt}, yet their representations are still rarely aligned with task-specific student models. Parallel work systematically evaluates the long-term memory capabilities of LLM agents\cite{maharana2024evaluating}, showing that models struggle to maintain consistent personalization over very long interactions. CogEvo-Edu's Cognitive Perception Layer instead maintains structured, confidence-weighted student profiles and couples their evolution tightly to the Knowledge Evolution and Meta-Control layers, so that profile changes directly influence retrieval, difficulty selection, and teaching strategies.

\subsection{Multi-Agent LLM Frameworks and Meta-Control}

Multi-agent LLM frameworks instantiate specialized agents that collaborate via dialogue and have been applied to planning, coding, and simulation\cite{wu2023autogen,li2023camel}. In educational AI, multi-agent paradigms are beginning to emerge as well; for example, the EducationQ framework\cite{shi2025educationq} uses multiple agents to simulate student--teacher dialogues, but typical coordination logic is still heuristic and optimized only for short episodes. At the same time, the LLM-as-a-Judge paradigm has become a practical way to evaluate open-ended responses, including in education, yet it is usually used purely as an external assessment tool\cite{li2025from}. CogEvo-Edu's Meta-Control Layer combines these ideas: it orchestrates multiple teaching agents through an inner-loop policy and uses a long-term objective, estimated by an LLM-as-a-Judge ensemble on DSP-EduBench, to update both this policy and the hyperparameters governing CPL and KEL, turning orchestration and memory/knowledge evolution into a unified meta-control problem.

\section{Methodology}

We introduce CogEvo-Edu, a cognitive evolution-based educational multi-agent system. In contrast to traditional static RAG-driven knowledge base systems, CogEvo-Edu possesses capabilities for cognitive awareness and knowledge evolution. The system comprises three core hierarchical layers: the Cognitive Perception Layer (CPL), the Knowledge Evolution Layer (KEL), and the Meta-Control Layer (MCL). Specifically, the CPL is responsible for constructing and maintaining dynamic student profiles; the KEL manages the full lifecycle of the knowledge base, including retrieval, expansion, and forgetting; and the MCL governs a dual-loop feedback mechanism to coordinate micro-level teaching strategies with macro-level system iteration.

\begin{figure*}[!b]
\centering
\includegraphics[width=0.9\textwidth]{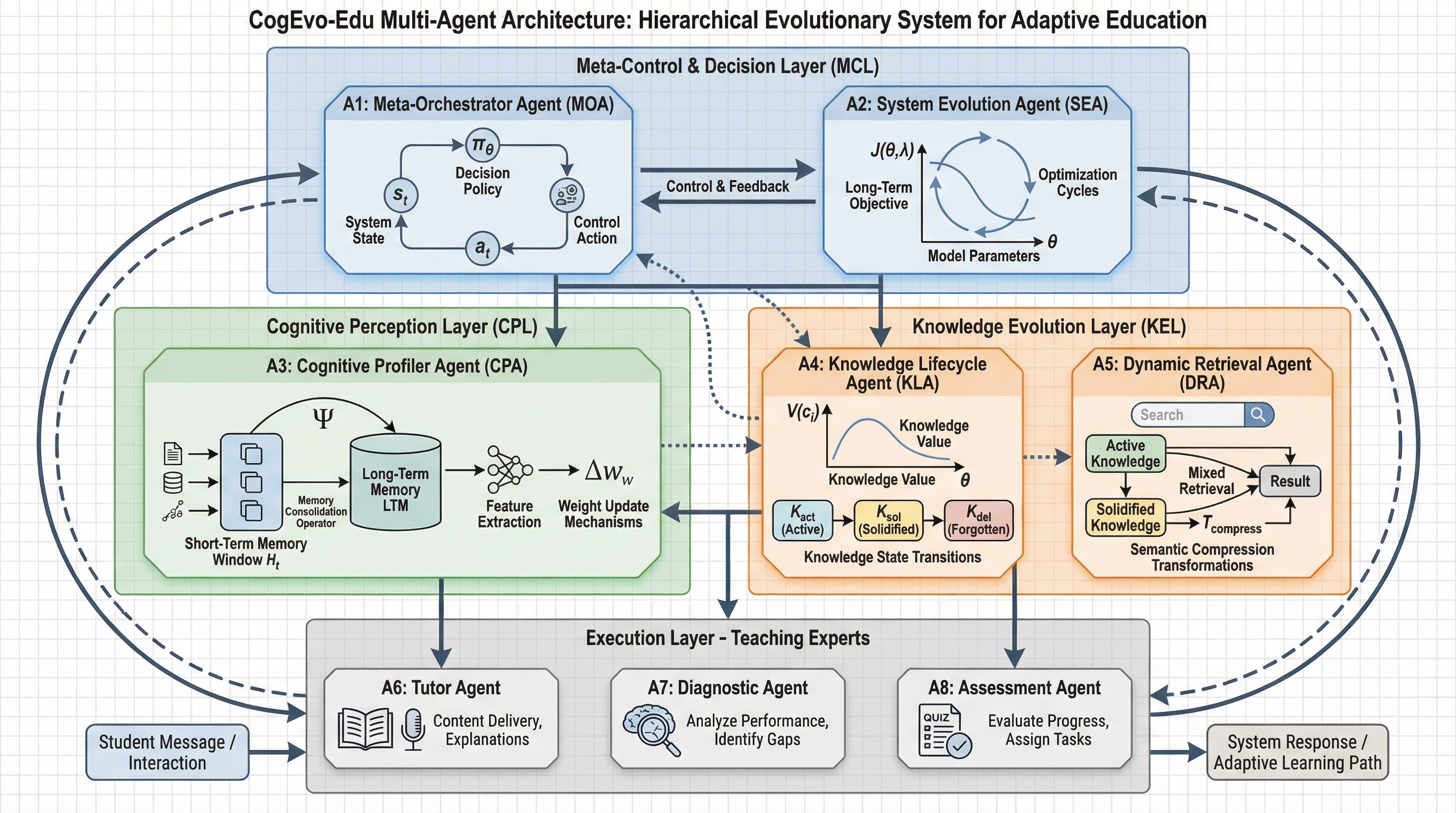}
\caption{CogEvo-Edu: Cognitive Evolution Educational Multi-Agent Collaborative System}
\label{fig:architecture}
\end{figure*}

\subsection{Cognitive Perception Layer (CPL)}

The core objective of the CPL is to achieve long-term tracking and precise modeling of student cognitive states under the constraints of finite context windows. To mitigate "catastrophic forgetting" in long conversations and the "Lost-in-the-Middle" phenomenon common in RAG systems, we design a hierarchical memory consolidation mechanism.

The CPL maintains two distinct memory storage spaces: Short-Term Sensory Memory and Long-Term Cognitive Memory. We model the profile construction problem as a sequential feature extraction and state update process.

We define the Short-Term Memory at time $t$ as the sequence $\mathcal{H}_t = \{(q_i, a_i)\}_{i=t-w}^{t}$, where $w$ represents the sliding window size. We define the Long-Term Memory as a set of structured features $\mathcal{P}_t = \{ (k_j, v_j, \omega_j) \}_{j=1}^M$, where $k$ is the feature key (e.g., "knowledge blind spot"), $v$ is the value, and $\omega \in [0,1]$ is the confidence weight. To address catastrophic forgetting, we design an incremental memory consolidation operator $\Psi$. When $\mathcal{H}_t$ saturates, $\Psi$ is activated to update the profile:

$$
\mathcal{P}_{t+1} = \Psi(\mathcal{P}_t, \mathcal{H}_t) = \mathcal{P}_t \oplus \text{LLM}_{extract}(\mathcal{H}_t)
$$

Here, $\oplus$ denotes the feature fusion operation. The specific fusion logic follows a semantic consistency update rule: for a new feature $f_{new}$ extracted from $\mathcal{H}_t$, we compute its semantic similarity $\text{sim}(f_{new}, f_{old})$ with existing features $f_{old} \in \mathcal{P}_t$. The update of the confidence weight $\omega$ follows a momentum mechanism:

$$
\omega_{new} =
\begin{cases}
\omega_{old} + \eta \cdot (1 - \omega_{old}), & \text{Reinforcement} \\
\omega_{old} - \eta \cdot \omega_{old}, & \text{Correction}
\end{cases}
$$

where $\eta$ is the learning rate, Reinforcement is triggered when $\text{sim}(f_{new}, f_{old}) > \tau_{match}$, whereas correction is activated once $f_{new}$ contradicts $f_{old}$. Through this formulation, CogEvo-Edu constructs high-fidelity user profiles with self-correction capabilities without increasing inference overhead.

Short-term memory primarily retains raw QA pairs from the most recent interaction turns to capture the student's current local performance; long-term memory records relatively stable features—such as knowledge mastery, problem-solving preferences, and error-prone patterns—in the form of structured entries. When sufficient new interactions accumulate in short-term memory, the system triggers a "consolidation" process: candidate features are first extracted from these dialogues and then compared against existing entries in long-term memory. If new evidence aligns with old conclusions, it is treated as a reinforcement event, increasing confidence; if a clear conflict arises, the original entry is weakened or corrected. By iterating this process, the system continuously distills stable profiles from local dialogues while progressively updating them upon the arrival of new evidence.

\subsection{Knowledge Evolution Layer (KEL)}

The core contribution of the KEL is the proposal of a spatiotemporal-aware value evaluation function, which drives the adaptive pruning and consolidation of the knowledge base.

Let the knowledge base $\mathcal{K}$ consist of a series of knowledge chunks, i.e., $\mathcal{K} = \{c_1, c_2, ..., c_N\}$. For any $c_i$, its dynamic value score $V(c_i)$ is defined as a weighted sum of three dimensions:

$$
V(c_i) = \alpha \cdot \underbrace{\frac{f(c_i)}{\max_j f(c_j)}}_{\text{Interaction Freq}} + \beta \cdot \underbrace{\exp\left(-\frac{\Delta t_i}{\tau_{decay}}\right)}_{\text{Time Decay}} + \gamma \cdot \underbrace{\mathcal{D}_{sem}(c_i)}_{\text{Semantic Density}}
$$

Here, Interaction Frequency $f(c_i)$ represents the number of times the chunk has been retrieved and adopted. Time Decay is determined by $\Delta t_i$, the time elapsed since the last access, with $\tau_{decay}$ being the half-life constant simulating the forgetting curve. Semantic Density $\mathcal{D}_{sem}(c_i)$ measures the importance of the chunk within the knowledge topology. We approximate this by calculating the average cosine similarity between $c_i$ and its $k$-nearest neighbors:
$$
\mathcal{D}_{sem}(c_i) = \frac{1}{k} \sum_{c_j \in \text{KNN}(c_i)} \cos(\mathbf{e}_i, \mathbf{e}_j)
$$
Based on the score $V(c_i)$, we define two decision thresholds $\theta_{solid}$ and $\theta_{forget}$ (where $\theta_{solid} > \theta_{forget}$), partitioning the knowledge base into three state sets:
$$
\left\{
\begin{array}{l}
\mathcal{K}_{act} = \{\, c_i \in \mathcal{K} \mid V(c_i) \ge \theta_{solid} \,\} \\[2pt]
\mathcal{K}_{sol} = \{\, c_i \in \mathcal{K} \mid \theta_{forget} \le V(c_i) < \theta_{solid} \,\} \\[2pt]
\mathcal{K}_{del} = \{\, c_i \in \mathcal{K} \mid V(c_i) < \theta_{forget} \,\}
\end{array}
\right.
$$
Elements in $\mathcal{K}_{act}$ retain their full vector indices. Elements in $\mathcal{K}_{del}$ undergo physical deletion. For elements in $\mathcal{K}_{sol}$, we introduce a semantic compression transformation $T_{compress}$: $c'_i \leftarrow \text{LLM}_{summ}(c_i)$, retaining only highly condensed abstract indices. This mechanism enables the system to achieve a theoretically optimal balance between storage cost $O(|\mathcal{K}|)$ and knowledge coverage $Recall@K$.

\subsection{Meta-Control Layer (MCL)}

The Meta-Control Layer sits above the CPL and KEL, responsible for unifying "how to teach, what to teach, and who teaches" into an optimizable decision problem. Unlike traditional systems reliant on handcrafted rules, the MCL conceptualizes the teaching process as a hierarchical sequential decision-making process: the bottom layer handles teaching action selection for single interactions, while the top layer manages cross-student, cross-session strategy and hyperparameter adaptation. This constitutes a dual closed-loop of Student Cognitive Evolution + System Strategy Evolution.

At each interaction timestep $(t)$, the state observed by the system is jointly provided by the CPL and KEL, formulated as:

$$
s_t = (\mathcal{P}_t, \mathcal{K}^{act}_t, x_t)
$$

where $\mathcal{P}_t$ is the structured student profile, $\mathcal{K}^{act}_t$ is the currently active knowledge subset, and $x_t$ is the current task question or concept information. Based on this state, the MCL selects a teaching action $a_t$, which includes: the leading agent role (Explanation, Diagnosis, Question Generation, Knowledge Reconstruction), teaching strategy (Direct instruction, Hinting, Socratic questioning, etc.), content difficulty, and retrieval configuration. The immediate system reward $r_t$ synthesizes learning performance and interaction cost to provide signals for subsequent strategy optimization.

Given fixed system parameters, the MCL implements micro-level teaching control via a parameterized policy $\pi_{\theta}(a_t \mid s_t)$. We treat the various functional agents in the system as a set of experts, and the MCL performs a soft orchestration of these experts based on the current student profile and knowledge value distribution. For instance, when the CPL identifies systemic misconceptions in a student, the MCL increases the weight of diagnostic and reflective agents; when the KEL indicates a high value density in a specific knowledge region, the MCL favors explanation and example generation strategies that can fully leverage that context. Simultaneously, the inner-loop policy adaptively adjusts problem difficulty based on mastery estimates from the profile and configures retrieval scope and compression levels according to KEL value scores, thereby achieving coordinated optimization between "what to teach" and "how to teach."

Sole reliance on a fixed inner-loop policy is insufficient to adapt to diverse student populations and long-term educational goals. Therefore, the MCL introduces an outer-loop meta-optimization mechanism over a longer time scale. Let $\theta$ be the inner-loop policy parameters and $\lambda$ be the set of hyperparameters controlling CPL and KEL behaviors (e.g., profile consolidation learning rate $\eta$, knowledge value function weights $\alpha, \beta, \gamma$, consolidation/forgetting thresholds $\theta_{solid}, \theta_{forget}$, and time decay constant $\tau_{decay}$). The system's long-term objective is defined as:

$$
J(\theta,\lambda) = \mathbb{E}\Big[\sum_{t} r_t\Big]
$$

representing the expected learning gain across multiple students and sessions. The MCL periodically aggregates interaction trajectories and estimates $J(\theta,\lambda)$ based on metrics such as knowledge mastery improvement, error pattern extinction, and retention rates, subsequently performing a joint update on $\theta$ and $\lambda$. Consequently, the intensity of profile updates in the CPL, the knowledge consolidation and forgetting strategies in the KEL, and the orchestration of multi-agents are no longer manually preset but are adaptively regulated in a data-driven manner oriented towards learning outcomes.

Under this dual-loop framework, the inner loop drives the gradual evolution of student profiles and the knowledge base alongside interactions under given parameters, manifesting as student cognitive evolution; the outer loop conversely adjusts the system's own control parameters and collaboration strategies on a cross-session scale, manifesting as system strategy evolution. The interplay between the two elevates CogEvo-Edu from a static RAG-driven system to a cognitive evolutionary educational multi-agent platform capable of continuous self-improvement.

\section{Experiments Setup}

\subsection{Dataset construction}

To evaluate CogEvo-Edu's cognitive evolution and multi-agent performance, we constructed DSP-EduBench, a vertical benchmark for Digital Signal Processing (DSP) tutoring. DSP deeply integrates abstract mathematical theory, physical intuition, and engineering code implementation—heterogeneous modalities and long logical chains that expose traditional RAG limitations in retrieval and cognitive alignment. Our dataset has three core layers: First, we curated heterogeneous sources (textbooks, mathematical derivations, MATLAB/Python code) to build a semantically connected knowledge base, testing KEL's ability to flexibly invoke resources across theory and code.

In order to transcend the limitations of static Q$\&$A testing and validate the profile construction capabilities of the Cognitive Perception Layer (CPL), we designed a dynamic interaction environment based on heterogeneous Simulated Student Profiles. We defined multiple typical profiles to cover diverse cognitive states: UserA, an intuitive novice lacking mathematical foundations and prone to definition amnesia; UserB, a learner with typical misconceptions who repeatedly falls into logical traps during problem-solving; and UserC, an advanced engineering student with solid foundations but a focus on implementation details and parameter tuning, among others. Based on these profiles, we generated long-range interaction scripts encompassing conceptual discrimination, fault diagnosis, and code debugging, forcing the system to maintain continuous tracking of specific student states across multi-turn dialogues. To ensure the objectivity of automated assessment, we annotated fine-grained Ground Truth for each interaction scenario. This includes not only standard mathematical solutions and factual truths but also checklists of key knowledge points that responses must cover, alongside annotations for the *ideal pedagogical strategy* tailored to the current dialogue context, thereby providing a reliable basis for subsequent automated evaluation.

\subsection{Evaluation Metrics}

we established an automated evaluation pipeline based on the LLM-as-a-Judge paradigm to quantify the performance of CogEvo-Edu across different architectural levels. The instructional model uniformly employs Qwen3-14B as the backbone, while the evaluation end adopts a model ensemble strategy, comprising an expert jury of GLM-4.5, DeepSeek-V3.1, and Qwen3-max. The jury independently scores each dialogue turn on a scale of 1 to 10 based on six key indicators across three dimensions, with the final score derived from the average to eliminate the bias of any single model.

\begin{figure}[htbp]
\centering
\includegraphics[width=\columnwidth]{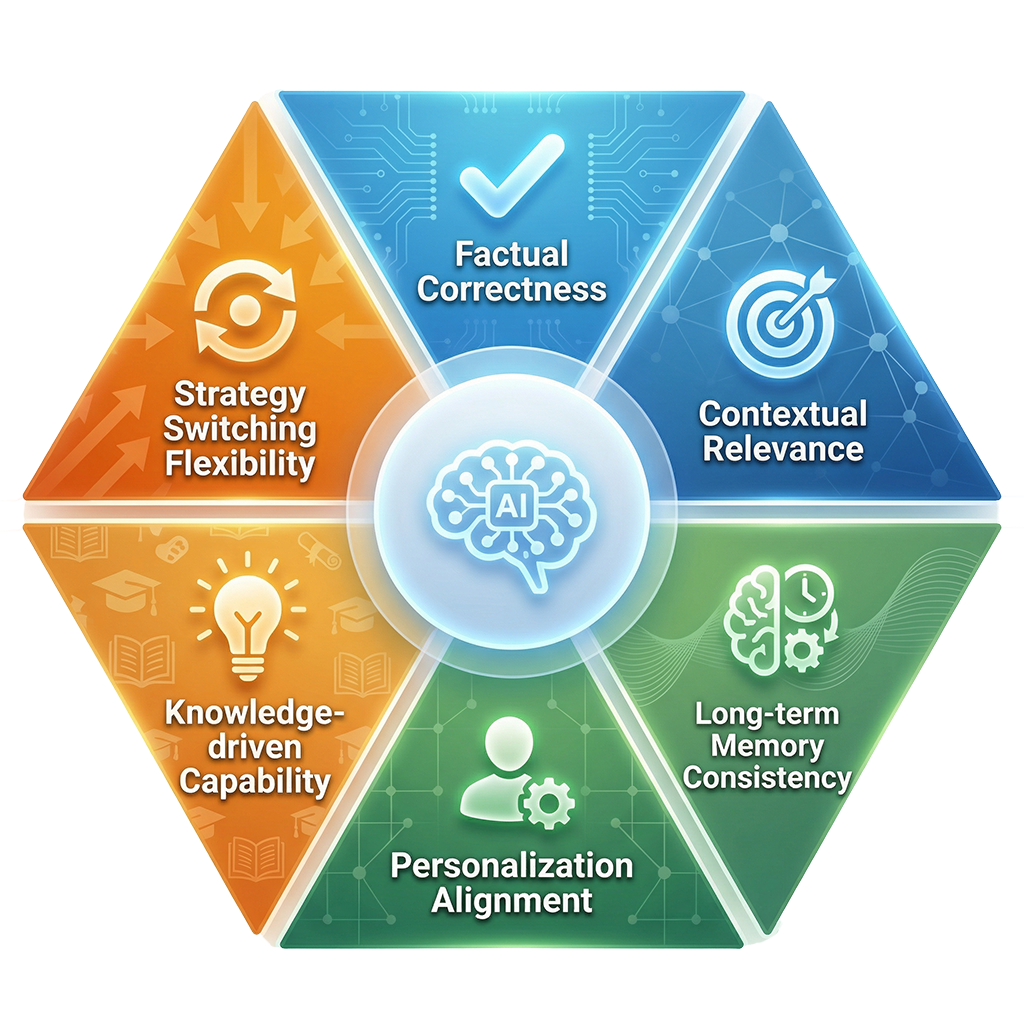}
\caption{Evaluation pipeline based on LLM-as-a-Judge principles.}
\label{fig:evaluation}
\end{figure}

To verify the effectiveness of the Knowledge Evolution Layer (KEL) in knowledge retrieval and denoising, we established the Knowledge Precision Dimension, comprising two indicators: \textbf{Factual Correctness} and \textbf{Contextual Relevance}. The former rigorously detects whether mathematical derivation errors or factual hallucinations exist in the model's responses. The latter focuses on examining whether retrieved content precisely addresses the student's current confusion and whether the system successfully filters out redundant information irrelevant to the current context via the value function, thereby avoiding the "retrieval piling" phenomenon common in traditional RAG systems.

Addressing the performance of the Cognitive Perception Layer (CPL) in overcoming catastrophic forgetting, we designed the Cognitive Coherence Dimension, evaluated through Long-term Memory Consistency and Personalization Alignment. \textbf{Long-term Memory Consistency} measures whether the system can still accurately recall a student's initial weaknesses and preferences during the later stages of long dialogues, ensuring no fragmentation in the instructional process. \textbf{Personalization Alignment} assesses whether instructional content dynamically adapts to the student's cognitive level; for instance, detecting whether the system automatically reduces formula density in favor of intuitive metaphors when facing novices, or provides in-depth code implementation details when engaging with advanced learners.

Finally, the evaluation system reflects the multi-agent orchestration and dual-loop control capabilities of the Meta-Control Layer (MCL) through the Pedagogical Strategy Dimension. This dimension includes two major indicators: Knowledge-driven Capability and Strategy Switching Flexibility. \textbf{Knowledge-driven Capability}, grounded in Bloom's Taxonomy, evaluates whether the system mechanically provides direct answers or guides students toward self-correction through step-by-step decomposition and Socratic questioning. \textbf{Strategy Switching Flexibility} examines whether the system, when facing instructional deadlocks, can break away from the current path and leverage the macro-regulation mechanism of the meta-control layer to flexibly switch from conceptual explanation to analogical demonstration or diagnostic testing.

\section{Experiments and Results}

We evaluate CogEvo-Edu on DSP-EduBench, our digital signal processing (DSP) tutoring benchmark with heterogeneous resources (textbooks, derivations, and code) and simulated student profiles spanning novices to advanced engineering learners. All models use Qwen3-14B as the instructional backbone and are assessed with an LLM-as-a-Judge ensemble (GLM-4.5, DeepSeek-V3.1, Qwen3-max), which assigns 1–10 scores along six dimensions: Factual Correctness, Contextual Relevance, Memory Consistency, Personalization Alignment, Knowledge Guidance, and Strategy Flexibility.

\subsection{Experimental Configurations}
We compare five configurations on identical multi-turn interaction scripts:(a) LLM Only: Vanilla Qwen3-14B without retrieval or memory.(b) Static RAG: Standard vector retrieval over the DSP knowledge base with fixed top-k and no student modeling.(c) Simple Memory: Sliding-window dialogue summary as a lightweight long-context memory.(d) Single Agent: A monolithic tutor equipped with CPL + KEL but using a fixed teaching policy (no meta-control).(e) CogEvo-Edu (Ours): Full system with hierarchical CPL, spatiotemporal KEL, and dual-loop MCL for multi-agent orchestration.

\begin{table*}[htbp]
	\centering
	\caption{Comparative Evaluation Results on DSP-EduBench (Score: 1–10)}
	\label{tab:comparative_results}
	\begin{tabular}{lcccccccc}
		\toprule
		\textbf{Experimental Setting} & \textbf{Factual} & \textbf{Contextual} & \textbf{Memory} & \textbf{Personalization} & \textbf{Knowledge} & \textbf{Strategy} & \textbf{Average} \\
		& \textbf{Correctness} & \textbf{Relevance} & \textbf{Consistency} & \textbf{Alignment} & \textbf{Guidance} & \textbf{Flexibility} & \\
		\midrule
		(a) LLM Only & 5.8 & 6.5 & 4.2 & 4.8 & 5.5 & 5.1 & 5.32 \\
		(b) Static RAG & 8.4 & 6.9 & 4.5 & 5.0 & 5.8 & 5.2 & 5.97 \\
		(c) Simple Memory & 6.1 & 6.7 & 7.6 & 6.8 & 6.0 & 5.5 & 6.45 \\
		(d) Single Agent & 7.9 & 7.5 & 6.2 & 5.8 & 6.2 & 4.9 & 6.42 \\
		\textbf{(e) CogEvo-Edu (Ours)} & \textbf{9.3} & \textbf{9.1} & \textbf{9.5} & \textbf{9.2} & \textbf{8.9} & \textbf{9.4} & \textbf{9.23} \\
		\bottomrule
	\end{tabular}
\end{table*}

\begin{figure}[htbp]
	\centering
	\includegraphics[width=\columnwidth]{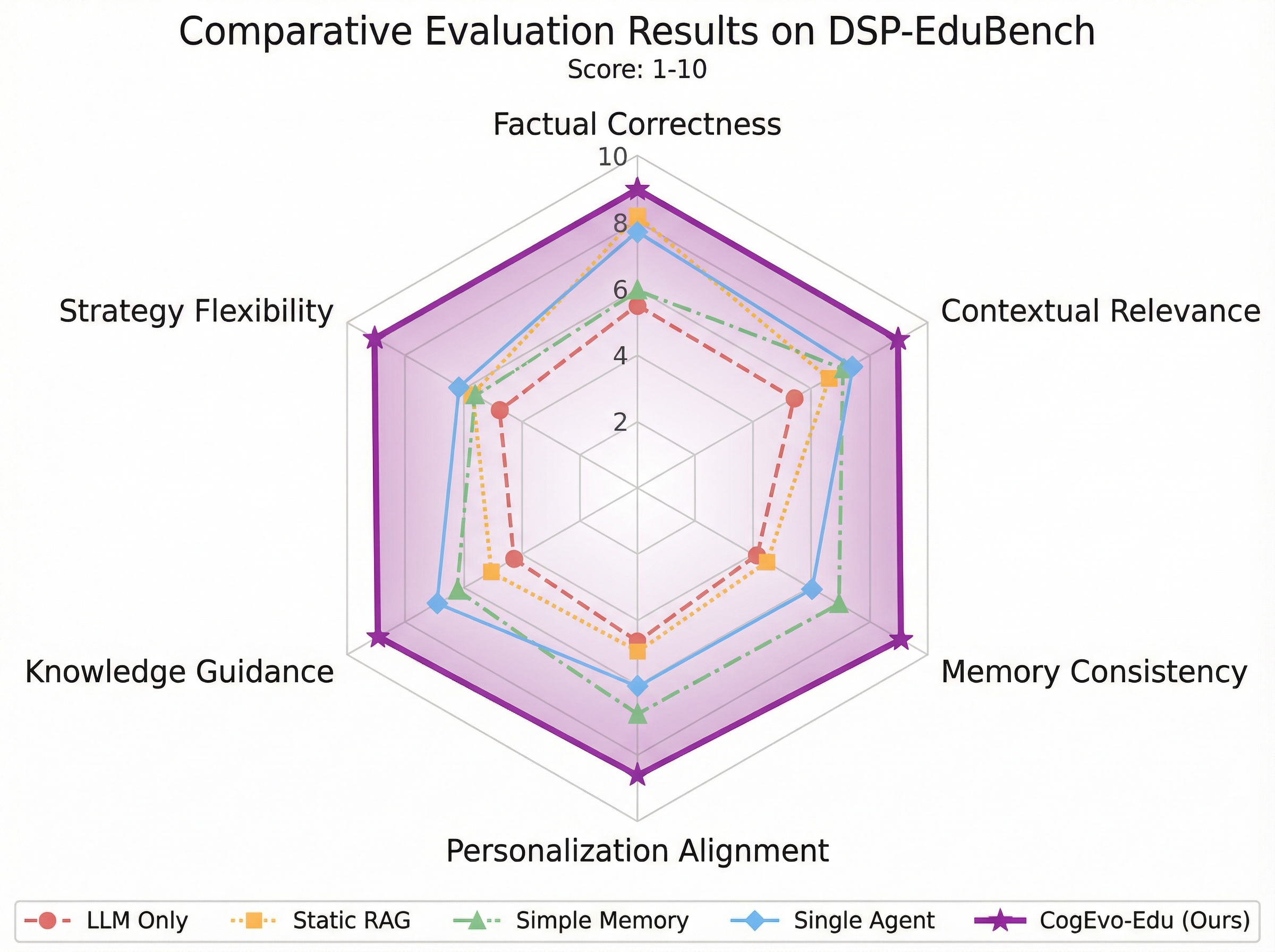}
	\caption{Comparative Evaluation Results Visualization Radar Chart}
	\label{fig:evaluation}
\end{figure}

\subsection{Overall Results}

Viewed along the three axes of knowledge precision, cognitive coherence, and pedagogical strategy, the baselines make the role of each component in CogEvo-Edu explicit. On the knowledge side, Static RAG already raises factual correctness substantially relative to LLM Only, but its gains in contextual relevance are modest, reflecting a “retrieval piling” effect where more documents are added without adequate filtering. In contrast, the value-guided KEL in CogEvo-Edu further improves both factual correctness and contextual relevance by dynamically pruning low-value content and semantically compressing redundant information, indicating that targeted retrieval rather than sheer retrieval volume is essential for high-precision DSP tutoring.

For cognitive coherence, Simple Memory enhances memory consistency by preserving recent dialogue, yet its unstructured summaries cap personalization alignment and make it difficult to maintain a stable, fine-grained model of the learner. Replacing this mechanism with CPL’s structured student profiles and confidence-weighted consolidation pushes both memory consistency and personalization alignment into the high-9 range, underscoring that explicit profile evolution, not generic long-context aggregation, is the key to long-horizon, student-aware interaction.

On the pedagogical axis, the Single Agent variant, which already incorporates CPL and KEL, attains reasonable knowledge guidance but remains rigid in strategy flexibility due to its fixed teaching policy. Introducing the MCL and multi-agent orchestration in CogEvo-Edu yields a sharp increase in both guidance quality and flexibility by adaptively selecting specialist agents and instructional modes as the session unfolds. Taken together, these comparisons show that the largest gains arise when retrieval, memory, and control are treated as a coupled cognitive evolution process, rather than as isolated enhancements to a monolithic tutor.

\section{Conclusion}
 DSP-EduBench demonstrates that CogEvo-Edu’s cognitive evolution design yields consistent, large-margin gains along all three axes of knowledge precision, cognitive coherence, and pedagogical strategy. The results suggest that future educational LLM systems should move beyond static RAG or monolithic tutoring toward explicitly modeled, jointly optimized evolution of student profiles, knowledge bases, and teaching policies.


\bibliographystyle{IEEEtran}
\bibliography{references}

\end{document}